\documentclass[runningheads]{llncs}
\makeatletter
\newcommand{\printfnsymbol}[1]{%
  \textsuperscript{\@fnsymbol{#1}}%
}
\makeatother

\usepackage{graphicx}

\usepackage{subcaption}
\usepackage{amsmath,algorithm,algpseudocode}
\makeatletter

\renewcommand{\Function}[2]{%
  \csname ALG@cmd@\ALG@L @Function\endcsname{#1}{#2}%
  \def\jayden@currentfunction{#1}%
}
\newcommand{\funclabel}[1]{%
  \@bsphack
  \protected@write\@auxout{}{%
    \string\newlabel{#1}{{\jayden@currentfunction}{\thepage}}%
  }%
  \@esphack
}
\makeatother

\usepackage{booktabs}
\renewcommand{\arraystretch}{1.2}
\newcommand{\ra}[1]{\renewcommand{\arraystretch}{#1}}
\renewcommand{\tabcolsep}{5pt}

\begin{document}
\title{NimbRo Logistics - Project KittingBot}
%
%
\author{Robin Rosche\inst{1}\thanks{Equal contribution} \and
Minh Triet Chau\inst{1}\printfnsymbol{1}}
\authorrunning{Robin and Minh Triet}
%
\institute{Rheinische Friedrich-Wilhelms-Universität Bonn, Bonn, Germany
\email{\{s6roros,s6michau\}@uni-bonn.de}}
\maketitle              
\begin{abstract}
Recovering the pose of an object from mere point clouds is often hindered by the lack of the information that they provide. In this lab, we address this problem by proposing a method that exploits the symmetry of objects as well as using pictures taken from a static camera of the same scene. We apply this approach to detects nuts in a table top scene that includes screws, nuts, washers and several placeholders for grasp planning.

\end{abstract}
\section{Introduction}
\begin{figure}[htbp]
    \centering
        \begin{subfigure}[b]{0.45\textwidth}
            \includegraphics[scale=0.25]{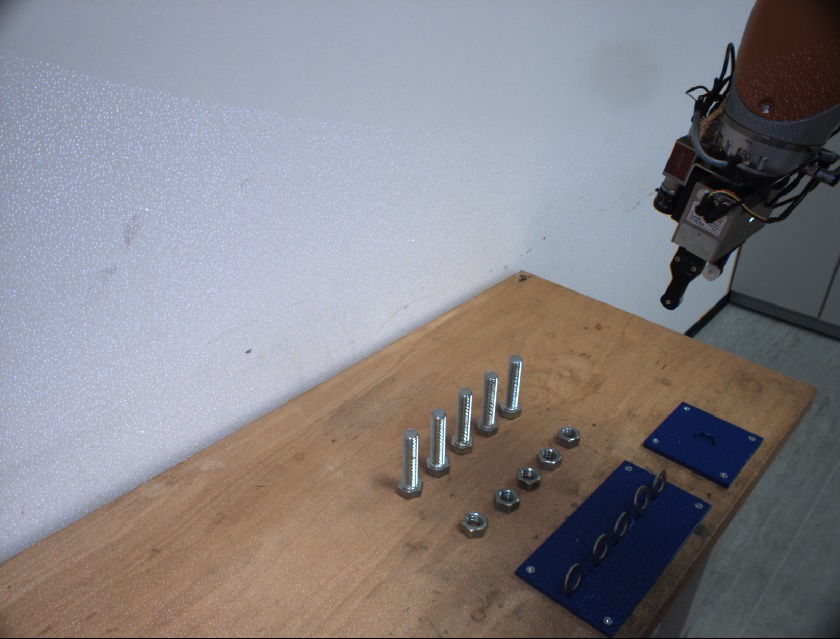}
            \caption{}
            \label{subfig:original_problem_left}
        \end{subfigure}
        \begin{subfigure}[b]{0.45\textwidth}
            \includegraphics[scale=0.25]{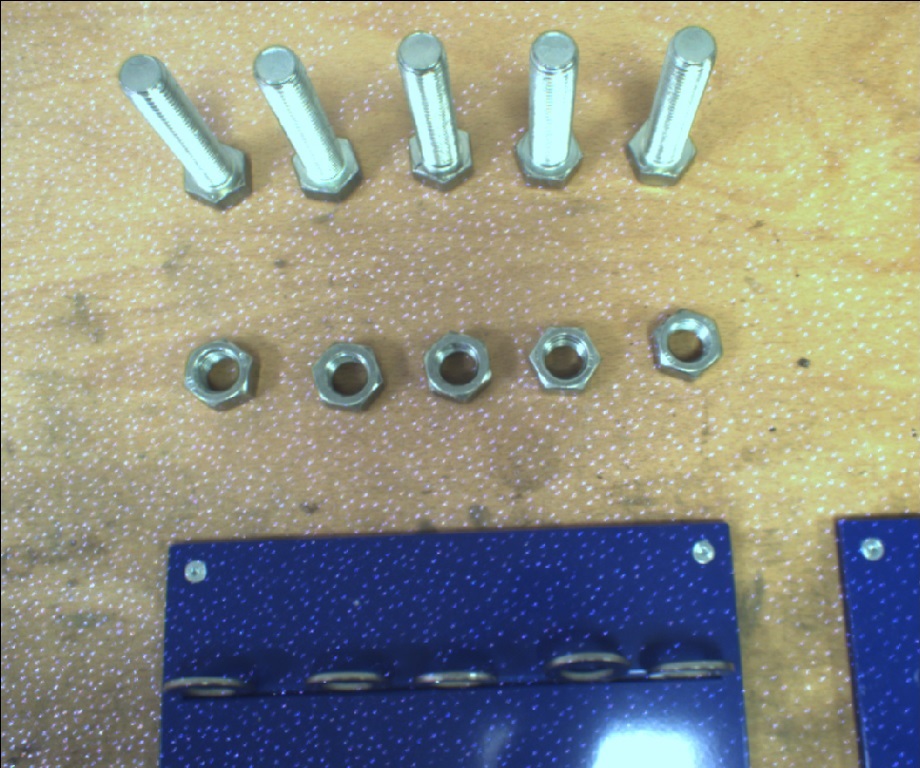}
            \caption{}
            \label{subfig:original_problem_right}
        \end{subfigure}
    \caption{The scenes where we try to detect poses of the nuts}
    \label{fig:original_problem}
\end{figure}

Given a table-top scene containing nuts, screws, washers and placeholders, we aim to tell the location and heading of the nuts (Figure \ref{fig:original_problem}). Since our only source of depth sensor is a static RGB-D camera, which is located at the left side of the table, a left nut tends to block the view of a right nut and so on. Therefore, we cannot assume that the point clouds will cover the whole nuts. Additionally, the quality of the point clouds that we receive is only useful in finding the centroid of the nut, not estimating the pose of the whole nut. Therefore, relying exclusively on the input point clouds to register the nut and estimate the pose is prone to errors.  In this report we present a solution by matching edges detected in that photo to projected 3D edges from rendered models of the nut. Note that in this problem, we only have two inputs dataset, as the setup of the challenge is fixed.

\section{Our Approach}
In this section we detail our method to solve the position matching of nuts and screws with a limited number of training samples, which is given through bagfiles. Since a bagfile gives point cloud messages separately at unknown time, the complete information about a nut segmentation is not available until the end of the bagfile. To avoid estimating a centroid with uncompleted point clouds, we only calculate the centroid after $N=11$ messages, $N$ is the number of point cloud messages in our bagfile.
\subsection{Data preprocessing}
\begin{figure}[htbp]
    \centering
    \includegraphics[scale=0.2]{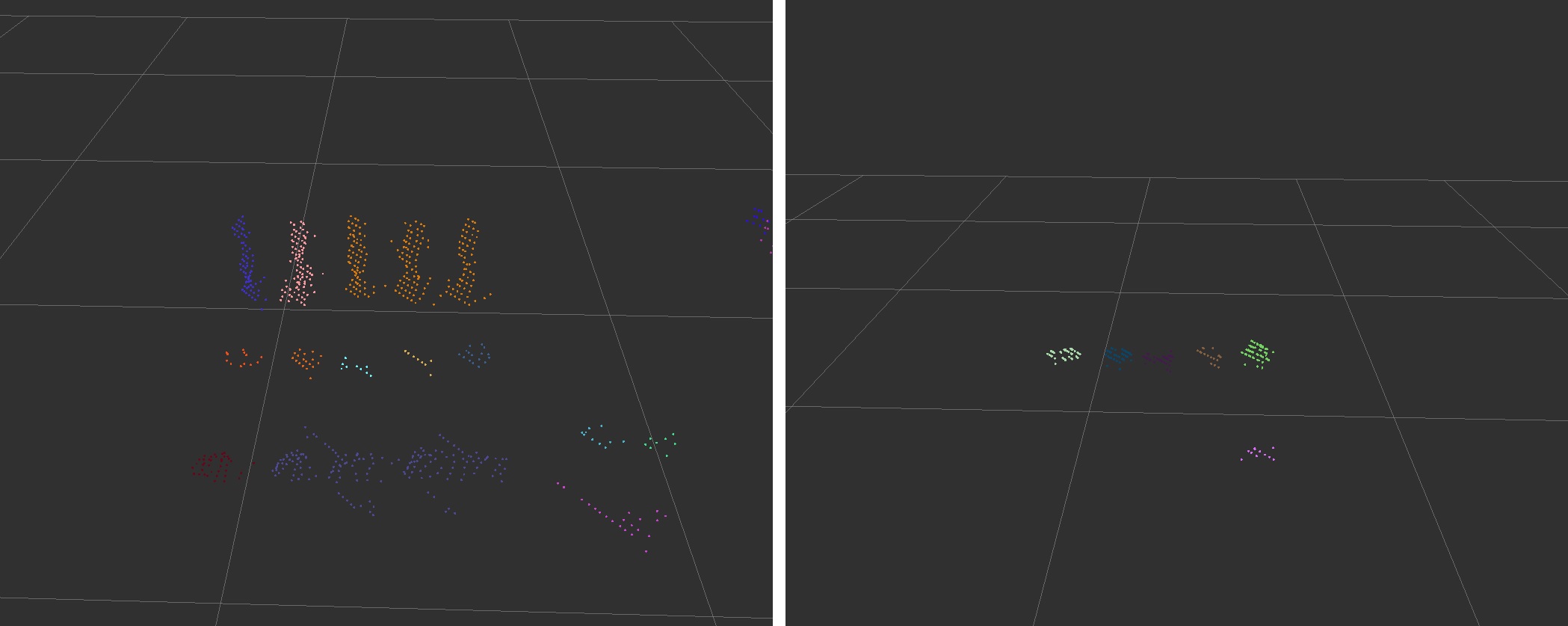}
    \caption{Left: Original point cloud data. Right: Filtered point cloud data.}
    \label{fig:before_after}
\end{figure}

First, from the bagfile, we only take in account the point clouds whose centroid lies between a lower and upper threshold on the table and satisfies a number of parameters that we have defined as follow: 
\paragraph{\texttt{min\_points}} Minimum number of points per cluster. This parameter serves to reduce noisy point clouds taken from the camera without discarding tiny bits of information. In our experiment, \texttt{min\_points} should not be larger than eight to ensure that we take in to account information of all nuts.
\paragraph{\texttt{min\_heights}} Minimal height of points with respect to the support plane in meters. \texttt{min\_heights} should be large enough to reduce unnecessary noise on the surface of the table. A too large value leads to reduced amount of points in the point cloud. In this experiment, we find the value of 0.008 gives the best result.
\paragraph{\texttt{max\_heights}} Maximal height of points with respect to the support plane in meters. We want to include the whole screws here for better filtering. Since the lower end of a screw has a similar shape to a nut, setting an incorrect value can make the program recognize the end of a screw as a nut. \texttt{max\_heights} should be a value larger than the height of a screw (0.12 meters) to ensure that we can differentiate between objects by their dimensions (Screws, washers, nuts).
\paragraph{\texttt{distance\_tolerance}} If the distance between two closest point in two point clouds is smaller than a value, these two point clouds can be merged together. There are cases that a screw comes in two separated point clouds, which leads to difficulties in distinguishing between nuts and screws, so it is desirable to merge these.
We then further compute a minimal bounding box around each point cloud to get its $x,y,z$ dimensions. We filter out point clouds by comparing their dimension size to a reference value which is taken from a real nut (Figure \ref{fig:before_after}). If the distance between two centroids is small enough, we merge these point clouds.

\subsection{Nuts pose detection}
\begin{figure}[htpb]
    \centering
    \includegraphics[scale=.6]{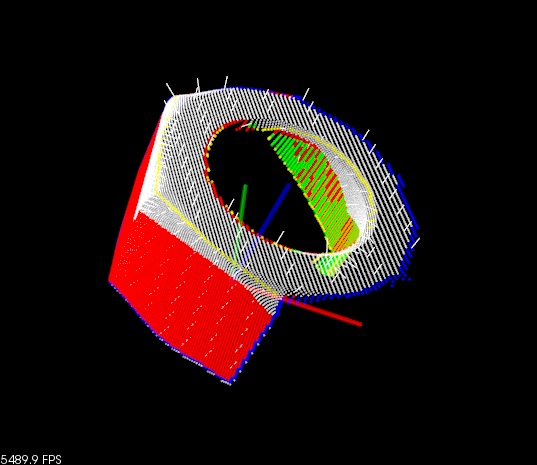}
    \caption{Modeled nut using the method described in \cite{choi}. The meaning of points is as follow. Blue points: Boundary edges. Green points: Occluding edges. Red points: Occluded edges. Yellow points: High curvature edges}
    \label{fig:nut}
\end{figure}

\begin{figure}
    \centering
    \includegraphics[scale=0.4]{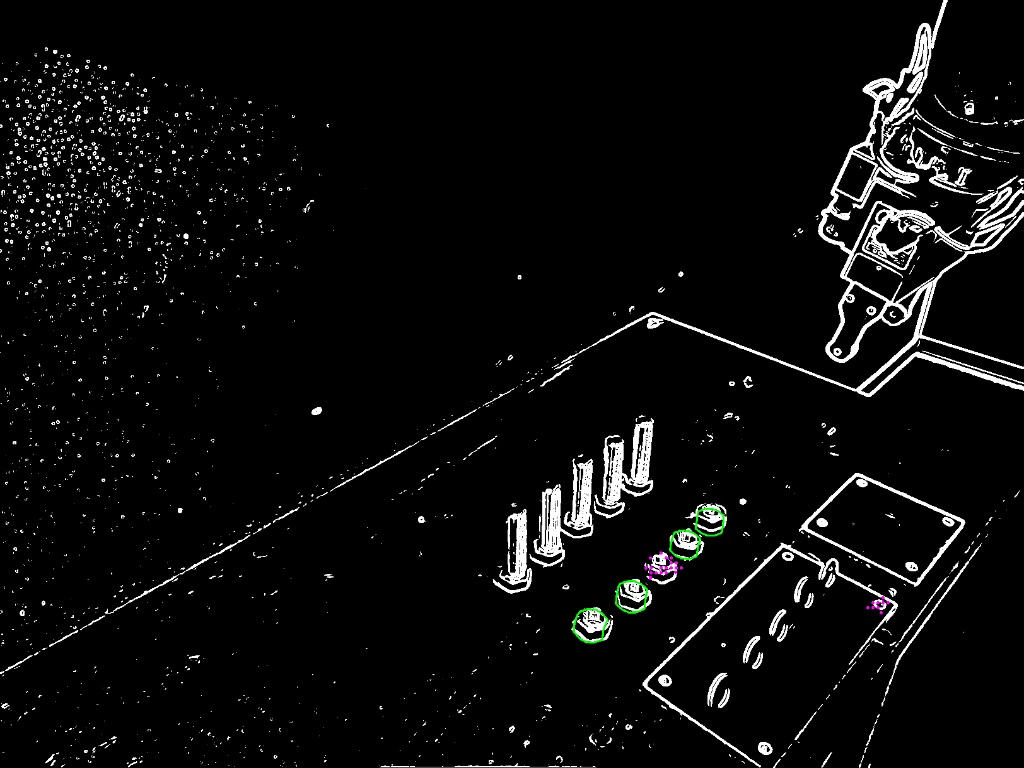}
    \caption{Projected 3d nuts to image}
    \label{fig:project_to_bin_img}
\end{figure}

Since our only data are from RGB-D caption of scene in Figure \ref{fig:original_problem}, we exploiting the symmetry of a nut and notice that the pose of a nut
is periodic in an interval of $60^\circ$. Therefore, from a given viewpoint and a centroid of a nut, we iterate through yaws from $[0^\circ$,$60^\circ] \approx [0,1.05]$ radian, with
an interval of 0.15 radian and render a nut with that yaw using a given CAD
model of a real nut. In this current yaw rotation setting, the maximum yaw
error is 0.075 radian (half of the rotation interval). 
We use the Scharr edge-detector and thresholding we generate a binary image of edges out of the camera photo (e.g. Figure \ref{fig:original_problem}). We then build a model of a nut (Figure \ref{fig:nut}) and project it to the binary image, varying the pose to maximize the score as in Algorithm \ref{alg:myalgo}. The result is in Figure \ref{fig:project_to_bin_img}

\begin{algorithm}
\caption{Edge ranking algorithm}
\label{alg:myalgo}
\begin{algorithmic}[1]
\Function{EdgeRanking}{$occluding\_edges, boundary\_edges, high\_curv\_edges$} \funclabel{alg:a} \label{alg:a-line}
    \For {edge \textbf{in} $occluding\_edges, boundary\_edges, high\_curv\_edges$}
        \For {point \textbf{in} edge}
            \State {score=(\Call{E}{$x \pm 1,y$}+\Call{E}{$x,y \pm 1$}+\Call{E}{$x \pm 1,y \pm 1$}+\Call{E}{$x,y$}) / 9}
            \If {score $\geq$ threshold}
                \State hits++
            \Else
                \State miss++
            \EndIf
        \EndFor
    \EndFor
    \State \Return hits / (hits+miss)
\EndFunction
\Statex
\Function{E}{$x,y$} \funclabel{alg:b}
    \If {there is an edge pixel in threshold image at location($x$,$y$)}
    \State\Return 1
    \Else
    \State\Return 0
    \EndIf
\EndFunction
\end{algorithmic}
\end{algorithm}

\section{Result}
Our result is depicted in Table \ref{tbl:left} and Table \ref{tbl:right}. As can be seen in the picture, the occluded edge (The red points in Figure \ref{fig:nut}) is not showed in the 2D image, so we remove them from consideration. We rank all remaining kinds of edges to find the best score.

\begin{table}
\centering
\ra{1.3}
\caption{Score of every nut from Figure \ref{subfig:original_problem_left}}
\label{tbl:left}
\begin{tabular}{@{}lccccc@{}}
\toprule 
 & Occluding edges & Boundary edges & High curvature edges & Score & Yaw \\
\midrule
Nut 1 & 0.770833 & \textbf{0.524126} & 0.527473 & 0.625914 & 0.15 \\
Nut 2 & 0.676259 & 0.419966 & 0.455598 & 0.641664 & 0.75 \\ 
Nut 3 & 0.224638 & 0.259197 & \textbf{0.554656} & 0.236202 & 0.3 \\
Nut 4 & 0.666667 & 0.430017 & 0.532075 & 0.605821 & 0.3 \\ 
Nut 5 & \textbf{0.877863} & 0.448336 & 0.506306 & \textbf{0.730623} & 0.9 \\
\bottomrule
\end{tabular}
\end{table}

\begin{table}
\centering
\ra{1.3}
\caption{Score of every nut from Figure \ref{subfig:original_problem_right}}
\label{tbl:right}
\begin{tabular}{@{}lccccc@{}}
\toprule 
 & Occluding edges & Boundary edges & High curvature edges & Score & Yaw \\
\midrule
Nut 1 & \textbf{0.95679}  & 0.528044  & \textbf{0.700269} & \textbf{0.728368} & 0.15\\
Nut 2 & 0.643243 & 0.415829  & 0.40479  & 0.487954 & 0.6 \\ 
Nut 3 & 0.50303  & 0.389349  & 0.454321 & 0.4489   & \textbf{0.75}\\
Nut 4 & 0.645349 & \textbf{0.535885}  & 0.669963 & 0.617066 & 0.3 \\ 
Nut 5 & 0.486188 & 0.506477  & 0.506767 & 0.49981  & 0.45 \\
\bottomrule
\end{tabular}
\end{table}

\begin{figure}
    \centering
    \includegraphics[scale=0.6]{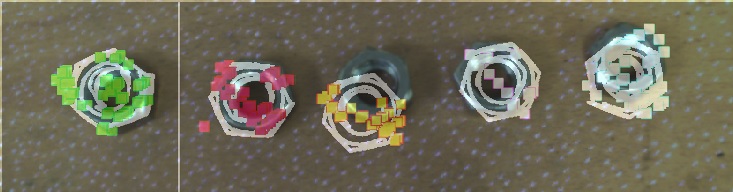}
    \caption{Bad distribution of points causes the middle nut to shift along the y-axis in the middle nut}
    \label{fig:bad_dist}
\end{figure}

As can be seen in Figure \ref{fig:bad_dist} the point cloud of the nut in the middle has a shifted centroid. Thus, the projected edges are also shifted and edge matching is unfeasible. Testing has revealed, that when the score is too low, rendering the nuts yield unreliable results. Due to a lack of testing data, it is still hard to determine an absolute lower bound for the score.

The distribution of the points in the point cloud is affected by the location of the RGB-D camera. In our lab, the camera is at the left side in front of the table, so the point clouds do not cover each nut completely. That makes the calculated centroid of a point cloud to differ greatly from the objects actual position. Therefore, we set the $z$-coordinate of every centroid to the known sum of the table and half of a nut’s height. That position of the camera also explains why the point cloud representation of the two left nuts leads to good results while problems arise in the case of other three nuts. The computed centroids of those nuts are shifted compared to their supposed position. This deviation propagates by projecting the nuts to the edge image. So altogether we have not achieved a reliable edge matching. 

\begin{figure}
    \centering
    \begin{subfigure}{0.45\textwidth}
        \includegraphics[scale=0.22]{fig7.jpg}
        \caption{}
        \label{subfig:last_sample_left}
    \end{subfigure}
    \begin{subfigure}{0.45\textwidth}
        \includegraphics[scale=0.22]{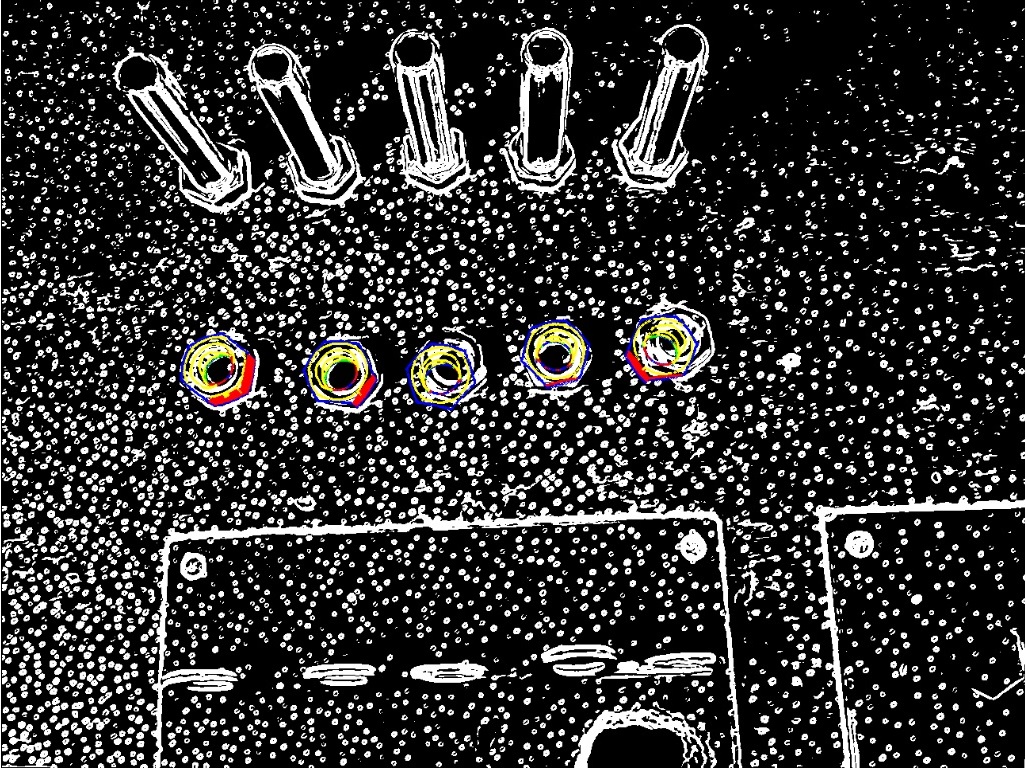}
        \caption{}
        \label{subfig:last_sample_right}
    \end{subfigure}
    \caption{The result of our method when applying to Figure \ref{subfig:original_problem_right}}
    \label{fig:last_sample}
\end{figure}

In case of Figure \ref{fig:last_sample}, the beam emitted by the infrared sensor causes a visible pattern on the camera image, which leads to a tremendous amount of noises. These noises causes falsely edge matching and render the edge image useless for ranking. Because of the noises in this image, Table \ref{tbl:right} shows a pretty good score value for the projected nuts that should not match with the edge image. Still, the perspective of Figure \ref{fig:last_sample} is desirable for detecting the correct pose of a nut because from this viewpoint one can see the complete contours of a nut.

\section{Conclusion}
To improve the results of our projection and edge matching approach, it is unavoidable to rely on additional sensor data. The point clouds are, naturally, important to find out the correct centroid of a nut. Furthermore the quality of the image which is taken for edge detection should be noiseless enough to not worsening the edge matching process. It takes a while to generate every position of the nut. We can improve that by storing all the prerendered nut in a file and access to them, rather than try to rerender everything from scratch. After that we can even lower the yaw-interval to get a more precise result.

Although there has been some improvements, our method is still sensitive to shift variant. That explains the poor result of the middle nut when its point cloud suggests a shifted centroid on the $y$-axis (Figure \ref{fig:bad_dist}). Another improvement would be using the RGB-D camera to take depth in- formation from several positions. Due to lack of additional depth information we could not investigate this any further.

\end{document}